\pdfoutput=1

\documentclass[11pt]{article}

\usepackage{acl}

\usepackage{times}
\usepackage{latexsym}
\usepackage{hyperref}

\usepackage[T1]{fontenc}

\usepackage[utf8]{inputenc}

\usepackage{microtype}
\usepackage{booktabs}
\usepackage{multicol}
\usepackage{multirow}
\usepackage{graphicx}
\usepackage{color}
\usepackage{array}

\definecolor{negative}{HTML}{CC0000}
\definecolor{positive}{HTML}{0000CC}
\definecolor{roller}{HTML}{246BCE}
\definecolor{chen}{HTML}{FF3800}
\definecolor{crook}{HTML}{12b637}

\newcommand{\POS}[1]{{\textcolor{positive}{#1}}}
\newcommand{\NEG}[1]{{\textcolor{negative}{\textit{#1}}}}

\def\mylinesep{\addlinespace[0.4em]}

\title{Teaching Models new APIs:\\ Domain-Agnostic Simulators for Task Oriented Dialogue}

\author{Moya Chen \\
  Facebook AI Research \\
  \texttt{mpchen@fb.com} \\\And
  Paul A. Crook \\
  Facebook Reality Labs \\
  \texttt{pacrook@fb.com} \\\And
  Stephen Roller \\
  Facebook AI Research \\
  \texttt{roller@fb.com} \\}

\begin{document}
\maketitle
\begin{abstract}

We demonstrate that large language models are able to simulate Task Oriented Dialogues in novel domains, provided only with an API implementation and a list of goals. We show these simulations can formulate online, automatic metrics that correlate well with human evaluations. Furthermore, by checking for whether the User's goals are met, we can use simulation to repeatedly generate training data and improve the quality of simulations themselves. With no human intervention or domain-specific training data, our simulations bootstrap end-to-end models which achieve a 37\% error reduction in previously unseen domains. By including as few as 32 domain-specific conversations, bootstrapped models can match the performance of a fully-supervised model with $10\times$ more data.
To our knowledge, this is the first time simulations have been shown to be effective at bootstrapping models without explicitly requiring any domain-specific training data, rule-engineering, or humans-in-the-loop.

\end{abstract}

\section{Introduction}

Virtual Assistants have become ubiquitous in modern life \cite{acharya2021alexa}. However, building these Task Oriented Dialogue (TOD) systems is laborious, requiring significant data collection and engineering resources to add support for a novel domain. As such, methods which can generalize, learn from limited examples, and require fewer engineering resources are highly desirable \cite{shi2019build,shah2018bootstrapping,acharya2021alexa}.

\begin{figure*}[t]
    \centering
    \includegraphics[width=0.8\linewidth]{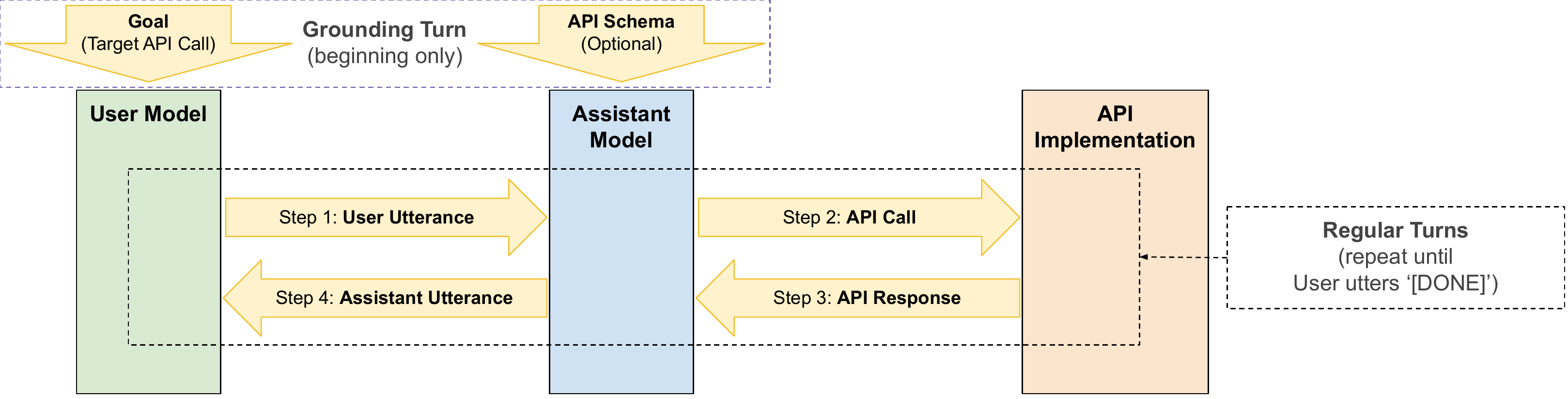}
    \caption{Illustration of our Simulation system. Initially, the User model is given a Goal API call and the Assistant is optionally provided an API Schema as grounding. Both models engage in dialogue until the User terminates the conversation. A dialogue is successful if the Assistant generates the correct API call by the end of the dialogue. In-arrows designate inputs to an entity; out-arrows designate what it generates.}
    \label{fig:arch}
\end{figure*}

To this end, works have previously identified User Simulators, wherein a model is used to emulate a human user in place of a real one, as a means of addressing these problems. User Simulators have been used to evaluate \cite{walker1997paradise,walker2000paradise,schatzmann2005quantitative} and improve Assistant models by providing additional training data \cite{shah2018bootstrapping,acharya2021alexa} and reward signals for Reinforcement Learning methods \cite{fazel2017learning,su2018d3q,shi2019build}. Typically, these User Simulators are either limited to enhancing existing domains \cite{fazel2017learning} or utilize specialized and manually engineered rules or templates for novel domains \cite{shah2018bootstrapping,shi2019build}. User Simulators have often required post-hoc human intervention to ensure quality \cite{shah2018bootstrapping}.

In the first part of this work, we show that modern Large Language Models \cite{radford2018improving,radford2018gpt2,lewis2020bart} are capable of reasonably generating dialogues when equipped with an API implementation and a desired User goal. We observe the quality of these dialogues increases with the quality of the models. Furthermore, we observe that simulation success is a strong discriminator of Assistant performance and dialogue quality.

In the second part of this work, we describe a method for bootstrapping User and Assistant models for previously unseen dialogue domains. We use Task Success, which can be automatically measured in fully synthetic dialogues, to discriminate between high- and low-quality dialogues. By adding successful dialogues back into the training set, retraining the model, and repeating this procedure, we bootstrap an Assistant model without the use of any domain-specific training data, hand-engineered rules, Natural Language templates, or humans-in-the-loop. Our methodology shows improvements in both zero-shot and full-shot settings.

Furthermore, we show that we can use Task Success as a method for automatically identifying the weakest skills of our model, and employ Active Learning \cite{tur2003active,olsson2009literature} to enhance performance. By additionally including as few as 32 domain-specific training examples, we can match the performance of a fully-supervised baseline provided with $10\times$ more data.

We open source our simulation infrastructure as part of the ParlAI framework \cite{miller2017parlai}.\footnote{\url{https://parl.ai/projects/tod_simulator/}}

\section{End-to-End TOD Simulators}
\label{sec:neuralusersim}

A high-level illustration of our simulation system is shown in Figure~\ref{fig:arch}. Our simulation system consists of three main components: a \textit{User} model, an \textit{Assistant} model, and an \textit{API Implementation}. A conversation consists of repeated turns between the User, the Assistant, and the API Implementation. Although traditional TOD systems model conversations with a combination of intent detection, belief state tracking, and policy \cite{jurafsky2009speech}, we employ a more modern setup \cite{rastogi2019towards} where the Assistant must both make  API calls at the right time and translate returned API responses into Natural Language utterances for the User. This formulation is particularly amenable to modern End-to-End (E2E) approaches based on pretrained Language Models \cite{ham2020end,peng2020soloist,hosseini2020simple}.

In order to guide the conversation, the User is given a \textit{Goal} as its first turn. This Goal consists of a complete API call (e.g.~intent, slot names, and slot values) serialized as a string. This is explicitly not shown to the Assistant model, to prevent the Assistant from seeing which API call it must make.
In each following turn, the User provides a natural language utterance to the Assistant. The Assistant optionally generates a serialized API call string and sends it to the API Implementation. If a successful API call is made, the API Implementation returns a serialized API response to the Assistant; a sentinel value is returned for any failed calls. The Assistant generates a natural language utterance back to User after receiving an API response.

A conversation continues in this repeated fashion, with turns of User utterance $\rightarrow$ Assistant API call $\rightarrow$ API Implementation response $\rightarrow$ Assistant utterance, until the User generates a `[DONE]' token or a maximum number of turns is exceeded. A conversation is said to be \textit{successful} if the Assistant generates the correct API call (i.e.~one equivalent to the Goal given to the User) before the end of the conversation. In order to have a successful conversation, we expect the User must ground its generations on the Goal, and the Assistant to ground its generations on the API response. We will later show that a simulation's \textit{Task Success Rate} (TSR), or its success averaged over a large number of goals, acts as a strong proxy for the quality of dialogues generated \cite{schatzmann2005quantitative}.

In addition to Goals being shown to the User agent, our system optionally allows the Assistant to be shown an \textit{API Schema} on the first turn. An API Schema consists of the signature of the Goal (e.g.~intent and slot names) without any realized values. As we will later see, Schemas (combined with Task Success) enable us to bootstrap models in unseen domains, but Schemas are \textit{not} used during any evaluations: doing so would leak the intent of the User and reduce the difficulty of the task.

\section{Related Work}

User Simulators have a long and successful history spanning many years, having been utilized widely for both evaluation \cite{walker1997paradise,schatzmann2005quantitative,ai2008user,jung2009datadriven,crook2017sequence}, and Assistant improvements \cite{li2016user,su2018d3q,shi2019build}. Their formulations have varied included Rule-based \cite{shi2019build}, Agenda-based \cite{schatzmann2007agenda,shah2018bootstrapping}, and End-to-End \cite{asri2016sequence,crook2017sequence,shi2019build} approaches.

A wide range of works have explored using User Simulators as a proxy for Assistant evaluation \cite{schatzmann2005quantitative} or predicting real user satisfaction \cite{walker2000paradise,ai2008user,jung2009datadriven, li2016user,crook2017sequence}. Performance of simulators are often measured either via Task Success Rate \cite{gur2018user,kreyssig2018neural} or by cross-examining them against a wide-variety of Assistant systems \cite{schatzmann2005quantitative}. Previously many have explored the use of Language Models (LMs) as User Simulators, but observed models had poor adherence to goals \cite{georgila2006usersim,crook2017sequence}. Accordingly, most simulators have been Agenda-based simulators, which follow templates and agendas to fulfill their goals \cite{schatzmann2007agenda,shah2018bootstrapping}. In contrast, we find that modern pretrained Language Models are able to ground on and follow goals very well (Section~\ref{sec:synthetic}).

Engaging User Simulators and Assistant systems in synthetically generated dialogues leads to a natural reward function, and many works have used simulators in order to optimize a Reinforcement Learning policy \cite{schatzmann2007agenda,fazel2017learning,peng2017composite,su2018d3q,gur2018user,kreyssig2018neural}. Such approaches are particularly used for optimizing the policy component of pipeline-based systems \cite{fazel2017learning}, and frequently rely on the use of Natural Language Generation (NLG) templates over dialogue acts \cite{fazel2017learning,shah2018bootstrapping,shi2019build,kreyssig2018neural, acharya2021alexa}. Our work instead utilizes fully lexicalized, E2E models for both the User and the Assistant models, without the need for agendas, dialogue acts, or NLG templates.

Most closely related to our work is that of \citet{shah2018bootstrapping} and \citet{acharya2021alexa}. Similar to our approach, both use User Simulators and Schemas to generate synthetic data and add them to a training dataset. However, \citet{shah2018bootstrapping} uses an additional stage of human-annotation in order to identify failed dialogues and paraphrases, while we show that Task Success can be a sufficient metric without Human involvement; we also utilize sampling to achieve diversity \cite{holtzman2020nucleus}. Furthermore, \citet{shah2018bootstrapping} perform only a single iteration of simulation, while we demonstrate improved performance from multiple iterations. \citet{acharya2021alexa} generalizes in a few-shot setting using a similar loop as ours, but rely on templates for NLG and human paraphrases. In contrast, we show we are able to gain performance improvements without dialogue acts or human paraphrasing, but we also show significant improvements via few-shot annotations using Active Learning.

End-to-End systems for TOD have had a surge in interest recently \cite{asri2016sequence,bordes2016learning,liu2018end,rastogi2019towards,ham2020end,hosseini2020simple,peng2020soloist,lin2020mintl}, owing to the success of utilizing pretrained models \cite{devlin2019bert,radford2018gpt2,lewis2020bart}. E2E models promise to lower the cost of annotation by replacing traditional pipeline models with text-in-text-out and adjacent external API calls \cite{rastogi2019towards,byrne2021tickettalk}. Compared to these works, we leverage pretrained models for User models in order to produce simulations, rather than only modeling Assistants. We also leverage Schema-based grounding techniques \cite{rastogi2019towards,balaraman2021domain}, which may be viewed as a form of in-context prompting methods \cite{brown2020language,schick2020small,wang2021towards,wei2021finetuned}.

\section{Evaluating Synthetic Dialogue Generation and Related Metrics}
\label{sec:synthetic}

We experiment to validate the appropriateness of our E2E TOD Simulator setup. We generate synthetic conversations with a variety of different model architectures and validate that traditional automated offline metrics, our TSR metric, and human evaluation  correlate positively and inline with general expectations. Note that our objective is to verify that modern end-to-end models are able to produce reasonable synthetic conversations using only goals, and not to obtain State-of-the-Art performance. While we expect more `powerful' model architectures to perform better than weaker ones, we primarily aim to validate the directional correlation of metrics across models.

\subsection{Experimental Setup}

\paragraph{Dataset}
\label{googleSgdDescripition}
We focus our efforts on the Google Schema Guided Dialogue (Google SGD) dataset \cite{rastogi2019towards}. Google SGD is a large TOD dataset with emphasis on zero-shot incorporation of new skills: models may make Service Calls (API Calls) which return responses. As part of the data, models may have access to \textit{API Schemas}, which contain call signatures of each of the possible services and intents. The held-out test set contains several services and intents which are not shown in the training data. In this section, we do \textit{not} employ the API Schemas. Instead, we train Assistant models that must memorize the underlying Schemas directly from the training data.

\paragraph{Models}
We experiment with four model architectures: LSTM \cite{hochreiter1997long}, LSTM with Attention \cite{bahdanau2014neural}, GPT2 \cite{radford2018gpt2}, and BART \cite{lewis2020bart} with R3F \cite{aghajanyan2020r4f}. We expect models later in this list to be more powerful. Our GPT2 implementation closely resembles the setup of SimpleTOD \cite{hosseini2020simple}, while our BART model roughly mirrors the implementation of MinTL \cite{lin2020mintl}.

\label{gsgd_setup} We fine-tune on Google SGD , using the original splits from \citet{rastogi2019towards}. For each model architecture type, we fine-tune separate User and Assistant models. We generate synthetic conversations with the setup described in Section \ref{sec:neuralusersim}. We extract goals from the validation dataset, leaving out conversations with multiple goals; each conversation is seeded with a single API call. We mock API Implementations via a lookup table with fully realized API calls as keys and corresponding API responses as values; this lookup table is populated directly from the dataset, and we return a sentinel failure value if the Assistant requests any API call not in the dataset. 

\paragraph{Evaluation Metrics}
We report two metrics for the Assistant: Joint Goal Accuracy (JGA)\footnote{We deviate from original Google SGD evaluation here and report JGA on the \textit{API calls} rather than the \textit{belief state}. An example receives a JGA score of 1 iff it generates the exact API call perfectly. Note that the majority of turns do not have API calls, so majority baseline is about 0.71.} and BLEU score. We additionally evaluate our simulation quality by Task Success Rate over the goals of the Valid and Test sets. In Google SGD, the Test set contains Out-of-Domain (OOD) examples that do not appear in the Train or Valid sets. As such, considering both Valid and Test gives us some estimate of OOD performance. While we aim to ensure that the models are well trained, e.g.~comparable to results in the literature, our main objective to examine the correlation between Task Success measured on synthetic data and established offline metrics.

 \paragraph{Human evaluation}
 We use ACUTE-Eval \cite{li2019acute} for human evaluation. ACUTE-Eval is a pairwise evaluation in which an annotator is shown two dialogues and asked a question about which they prefer. We ask annotators the question ``Which Assistant would you rather use yourself?'' As recommended by \citet{li2019acute}, we use a manually-curated control pair comparing an artificially repetitive dialogue with a gold dialogue from Google SGD; annotators who failed to identify the gold dialogue were removed. We select a random subset of 400 goals derived from the Valid set of Google SGD and use the same model architecture for both the User and Assistant to generate synthetic conversations. We only present User and Assistant utterance turns to annotators (hiding any API calls). We collect pairwise annotations between each model architecture described above, as well as the gold dialogues from the original dataset (Human). Annotators were presented with conversations with the same goal when comparing model-generated conversations. We measure the fraction of times each model architecture (or Human) was preferred in its pairwise match up, and compare all possible pairs of model architectures. An image of the annotator UI is included in Appendix~\ref{sec:screenshot}.

\subsection{Results}
\label{sec:synthdataresults}

\paragraph{Automatic evaluations}

\begin{table}[t]
    \centering
    \begin{small}
    \begin{tabular}{lrrrrr}
    \toprule
    & {\bf User} & \multicolumn{2}{c}{{\bf Assistant}} & \multicolumn{2}{c}{{\bf Simul. TSR}}\\
    \cmidrule(lr){2-2} \cmidrule(lr){3-4} \cmidrule{5-6}
    {\bf Model} & BLEU & JGA & BLEU & Valid & Test\\
    \midrule
    LSTM      & .058 & .777 & .123 & .042 & .042\\
    LSTM+Attn & .078 & .833 & .183 & .302 & .169\\
    GPT2      & .093 & .869 & .223 & .474 & .307\\
    BART      & .116 & .897 & .252 & .583 & .352\\
    \bottomrule
    \end{tabular}
    \end{small}
    \caption{Fully automatic metrics of different modeling approaches tested on the original Google SGD split.
    Simulation Task Success Rate (TSR) increases along with offline metrics, but shows
    greater discrimination in magnitude than offline metrics}.
    \label{tab:automatic_metrics}
\end{table}

Results are shown in Table~\ref{tab:automatic_metrics}. We find that performance of all metrics improves monotonically in the direction we expect: more modern models with better pre-training and regularization outperform others. We find that the magnitude of improvements from better modeling is more visible using either measure of TSR compared to using more traditional offline metrics

\paragraph{Human Evaluation}

\begin{figure}[t]
\centering
\begin{tabular}{lr|rrrrr}
\multicolumn{2}{c}{} & \multicolumn{5}{c}{Win \%}\\
\multicolumn{2}{c}{} & L & A & G & B & H\\
\cmidrule(lr){3-7}
\parbox[t]{2mm}{\multirow{4}{*}{\rotatebox[origin=c]{90}{Lose \%}}}
 & \underline{L}STM      &              & \NEG{.48}    & \POS{.55}    & \POS{.57}    & \POS{\bf.80}\\
 & \underline{A}ttn      & \NEG{.52}    &              & \POS{\bf.60} & \POS{\bf.63} & \POS{\bf.83}\\
 & \underline{G}PT2      & \POS{.45}    & \POS{\bf.40} &              & \POS{.58}    & \POS{\bf.81}\\
 & \underline{B}ART      & \POS{.43}    & \POS{\bf.37} & \POS{.42}    &              & \POS{\bf.75}\\
 & \underline{H}uman     & \POS{\bf.20} & \POS{\bf.17} & \POS{\bf.19} & \POS{\bf.25} &             \\
\end{tabular}
\caption{\textbf{Pairwise Human Evaluations of simulations} by the different models, along with Human conversations.
Entries marked in blue agree with automatic metrics, while those in red italics disagree with
the automatic metrics. Bold numbers indicate statistical significance ($p < .05$, binomial test).
LSTM models hallucinates realistic dialogues that ignore their goal.}
\label{fig:acute_metrics}
\end{figure}

The results of our human evaluation are shown in Figure~\ref{fig:acute_metrics}. We label the scores depending on whether they agree or disagree with our expectations from automatic metrics. We find only the LSTM v. LSTM-Attn pair disagrees with our expectations, while all other pairwise evaluations agree with our expectations. We also see that Humans are greatly preferred over all the simulations, indicating none of our simulations are at human-level performance. However, we additionally note that preference for gold data roughly decreases as the quality of the system improves.

We manually inspect training logs and reasons for human preferences. In the UI annotators were asked to provide commentary on their selections. As would be expected, successfully helping the User was often a provided reason for preferring one Assistant over another; conciseness, naturalness, and brevity were also mentioned.  Of particular note, we observe that the LSTM model generated only a few unique dialogues, and generally completely ignored its goal; though it had very low TSR, LSTM utterances would declare successful task completion. As a result, the LSTM essentially `hallucinates its way to success' since human annotators do not see the goals. On the other hand, the GPT2 and BART models ground strongly on the goals, and generate plausible dialogues for a given goal. In particular, we observe that due to redundancy in slot names across domains, stronger models are able to successfully perform some simulations of completely unseen domains. This ability to generalize is tested rigorously in the next section.

\section{Bootstrapping Novel Domains}
\label{sec:bootstrapping}

In the previous section, we demonstrated that User models can be adequately grounded on unseen goals to guide our simulations at generating plausible synthetic dialogues, possibly even on unseen domains. In this section, we consider whether this synthetic data can be used to \textit{bootstrap} models on completely novel domains.

At a high level, our approach depends on generating synthetic data, filtering the synthetic dialogues using Task Success, and re-training the simulation models while incorporating the synthetic dialogue. This process may be repeated for multiple iterations to form a feedback loop. 

\begin{table}[t]
    \centering
    \begin{small}
    \begin{tabular}{l}
    \toprule
    \textbf{Pretraining Data}\\
    \midrule
    Google SGD$^*$ \cite{rastogi2019towards}\\
    \mylinesep
    MultiWoz \cite{budzianowski2018multiwoz} \\
    MSR-E2E \cite{li2018microsoft}\\
    MetaLWoZ \cite{lee2019metalwoz} \\
    Taskmaster-1 \cite{byrne2019taskmaster1}\\
    Taskmaster-2 \cite{byrne2019taskmaster2} \\
    TicketTalk \cite{byrne2021tickettalk}\\
    MultiDoGo \cite{peskov2019multidogo}\\
    \bottomrule
    \end{tabular}
    \end{small}
    \caption{\textbf{Pretraining datasets.} For Google SGD, we use new splits described in Section~\ref{sec:bootstrapexp}.}
    \label{tab:pretraining}
\end{table}

\begin{table}[t]
    \centering
    \begin{small}
    \begin{tabular}{lrr}
    \toprule
    Fold                & No. Diag. & No. Domains\\
    \midrule
    Pretraining (train) &   119,677  &   29  \\
    \mylinesep
    In-Domain train     &   13,888    &    16 \\
    In-Domain valid     &   1,966     &    16 \\
    In-Domain test      &    3,132    &    16 \\
    \mylinesep
    Out-of-domain train &    2,303   &    4  \\
    Out-of-domain valid &    768     &    4  \\
    Out-of-domain test  &    768     &    4  \\
    \bottomrule
    \end{tabular}
    \end{small}
    \caption{\textbf{Dataset statistics} Datasets used for simulator pre-training and evaluation. In-Domain and Out-of-Domain refer to new splits described in \ref{gsgd_setup}. Pretraining statistics include those of In-Domain Google SGD. The Out-of-Domain train fold is used to sample goals and train baselines, and not for pretraining.}
    \label{tab:dataset_stats}
\end{table}

\subsection{Experimental Setup}
\label{sec:bootstrapexp}

\paragraph{Pretraining \& Data setup} Following the work of Soloist \cite{peng2020soloist} and TOD-BERT \cite{wu2020todbert}, we pretrain a
BART model on a large number of open source Task Oriented Dialogue
datasets. In early experimentation, we found pretraining improved zero-shot Out-of-Domain JGA performance by about 6 absolute points, and initialize with this pretrained model for all baselines and proposed models.
A complete list of the datasets used is shown in Table~\ref{tab:pretraining}. For brevity, full descriptions of each dataset and relevant preprocessing may be found in Appendix~\ref{sec:appdatasets}. 

To make sure that our out of domain bootstrapping of Google SGD is truly out of domain relative to pretraining, we build two custom splits for Google SGD. We analyze domains present across all of our datasets and select four holdout domains that are unique to Google SGD. We denote all conversations that use any of these holdout domains as part of the \emph{Out-of-Domain} split and all remaining conversations to be the \emph{In-Domain} split. Note that as Google SGD is a dataset which has multiple domains in a given conversation, some non-holdout appear as part of conversations in the Out-of-Domain split. Only In-Domain Google SGD is used for pretraining. All domains are listed in the Appendix.
Final statistics of pretraining and evaluation data are provided in Table~\ref{tab:dataset_stats}.

\paragraph{Bootstrapping Procedure}
We use a \emph{Schema-Aware} models in order to generate synthetic training data, as find that using schema aware models is necessary for zero-shot and few-shot domain generalization. We use goals from the Train split of Out-of-Domain Google SGD, to generating grounding for synthetic training data. In order to increase data diversity and prevent overfitting, we use Nucleus generation (\citeauthor{holtzman2020nucleus}, \citeyear{holtzman2020nucleus}; $p = 0.9$). We generate 20 synthetic conversations for a given goal and retain only successful conversations. These successful conversations make up the synthetic data that we use for fine-tuning, with 10\% withheld and used for model selection.

We fine-tune both Schema-Aware (for data generation) and Schema-Agnostic (for evaluation) versions of our models on this synthetic data. Recall that Schema-Aware models have the User intent leaked to them, and therefore only Schema-Agnostic models may be used for evaluation. Models are fine-tuned incrementally -- the best model from the previous iteration acts as initialization for the next iteration -- and synthetic data is accumulated across iterations. During early experimentation, we found that fine-tuning a single model on both User and Assistant roles generally performed better than fine-tuning separate models and use this multitask setup for all of our experiments. Additionally, we find that multitasking on In-Domain data alongside synthetic data helps prevent overfitting, and include it in all experimental conditions.

\paragraph{Experimental Comparisons}
We perform multiple experimental comparisons for models with synthetic data.

In our first experiment, we compare the performance of Schema-Aware and Schema-Agnostic models. This is to demonstrate that providing Schemas boosts performance and raises the Task Success Rate, enabling generation feedback loops.

In our second experiment, we consider how simulations may be used to bootstrap models in a \textit{Zero-shot setting}. In these experiments, we only provide Out-of-Domain information via goals and completely synthetic data. To show the improvement provided by simulations, we compare primarily against the Base pretrained model, which has never seen any Out-of-Domain information. To contextualize the result, we also provide results for a Fully-Supervised model upper baseline, which was fine-tuned directly on all available Out-of-Domain data. 

As models `in the wild' would not have access to Schemas (i.e.~they must detect user intent), we only evaluate the Schema-agnostic versions, and report offline metrics of Out-of-Domain JGA and Assistant BLEU-4, as well as the online metric TSR. We also report offline In-Domain JGA to ensure the use of synthetic data does not harm existing knowledge in the model.

In our third experiment, we consider how simulations enable a form of \textit{Active Learning}. In these setups, we intentionally add \textit{domain-specific} training data. At each iteration, we evaluate performance of the model across all the goals, and identify the 8 Schemas with the lowest overall performance. We select 8 conversations from the Out-of-Domain training set matching these goals, and add them into the training data in the next iteration (in addition to the synthetic data). We also select 8 conversations from the validation set and add them to the validation data. This method can be seen as a form of \textit{Active Learning}, where model performance by goals is used to guide data collection schemes at a lower total cost. As a comparison point, we evaluate models trained with an equal number of \textit{randomly} chosen samples, which demonstrate the performance of few-shot modeling without the use of simulation methodologies. To contextualize performance, we also compare to a model which uses $10\times$ more few-shot samples, and the fully supervised model.

We analyze these different model conditions in more granularity in Section~\ref{sec:deeperAnalysis}.

In our final experiment, we ensure we have not inadvertently overfit on the validation goals through accidental leakage. We test all the above models on the held-out test set, which contains entirely unseen goals and conversations. Additionally, we evaluate whether our bootstrap procedure can be used in data-rich environments by applying it on top of a fully-supervised model.

\begin{figure*}[t]
    \centering
    \includegraphics[width=\linewidth]{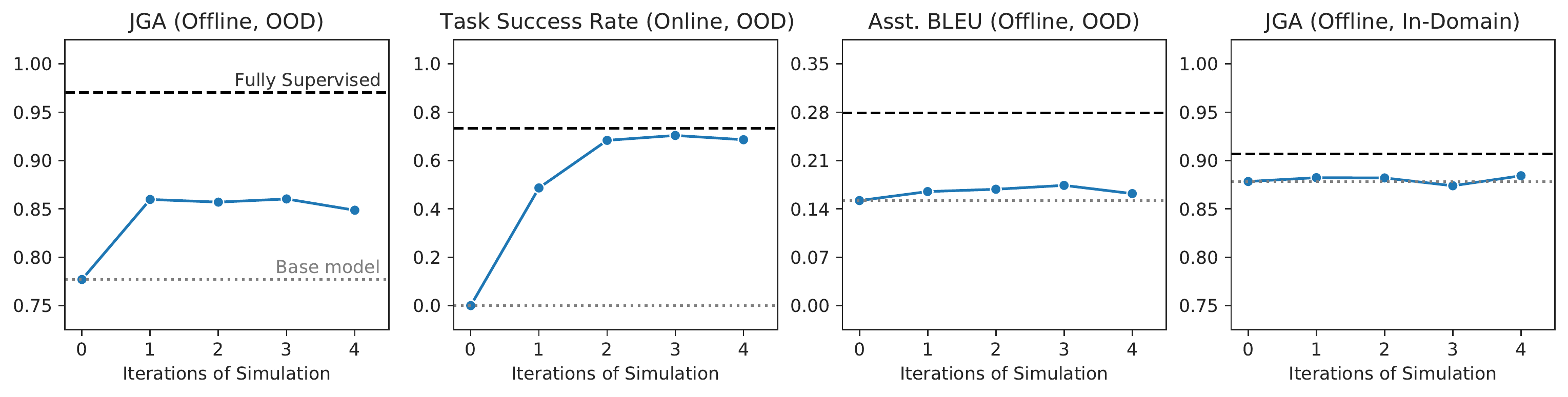}
    \caption{\textbf{Bootstrapped Performance with no Domain-specific data on validation data.} Out-of-Domain JGA and synthetic TSR both improve dramatically through only the use of synthetically generated data. Assistant BLEU and In-Domain JGA are unaffected by the synthetic data.}
    \label{fig:bootstrap_perf}
\end{figure*}

\subsection{Results}

\paragraph{Use of Schemas} Results of our first experiment are shown in Table~\ref{tab:schemaaware}. Across both In-Domain and Out-of-Domain, we see a substantial rise in performance in Schema-Aware models. This is unsurprising, as providing Schemas essentially cheats, allowing the model to bypass Intent detection. However, providing Schemas has a dramatic effect on Out-of-Domain performance, boosting JGA well above baseline performance, and enabling a non-zero Task Success Rate. Such successful conversations form the basis of our synthetic data during bootstrapping, and thus critical to our methodology.

\begin{table}[t]
    \centering
    \begin{small}
    \begin{tabular}{lrrrr}
    \toprule
    & \multicolumn{2}{c}{{\bf In-Domain}} & \multicolumn{2}{c}{{\bf Out-of-Dom}}\\
    \cmidrule(lr){2-3} \cmidrule{4-5}
    {\bf Model}     & JGA & TSR & JGA & TSR\\
    \midrule
    Schema-Agnostic   &  .878 & .292 & .777 & .000\\
    Schema-Aware      &  .960 & .839 & .880 & .369\\
    \bottomrule
    \end{tabular}
    \end{small}
    \caption{Comparing BART models with and without access to Schemas on Valid metrics. Schemas help guide the Assistant model to making the correct API calls in novel domains,
    as shown in TSR.}
    \label{tab:schemaaware}
\end{table}

\paragraph{Zero-shot}
Results for our Zero-shot experiments are shown in Figure~\ref{fig:bootstrap_perf}, with additional metrics provided in the Appendix~\ref{sec:appextraresults}. Overall, we find that Out-of-Domain JGA performance goes up by a total of 8.3 absolute points, or about a 37\% reduction in total errors, despite having no access to domain-specific data. However, JGA plateaus after just one iteration of simulation training, and further iterations provide marginal negative value.

On the other hand, TSR continues to improve for 3 iterations before eventually plateauing at approximately the level of the Fully-Supervised model; this suggests that JGA and TSR are no longer coupled, and that our bootstrapping procedure primarily optimizes its selection criteria: Task Success. To ensure the model was not simply making random API guesses to maximize TSR, we counted the number of API calls in simulation, and found the model converged on approximately 1 call per dialogue, matching the desired distribution.

We also find that Assistant BLEU does not change significantly compared to baseline, suggesting that improvements to Task Success do not translate to improvements in Natural Language Generation. Finally, we see that In-Domain JGA remains unchanged relative to the baseline, demonstrating that the addition of synthetic data does not come at the cost of performance in other domains.

\begin{figure}[t]
    \centering
    \includegraphics[width=\linewidth]{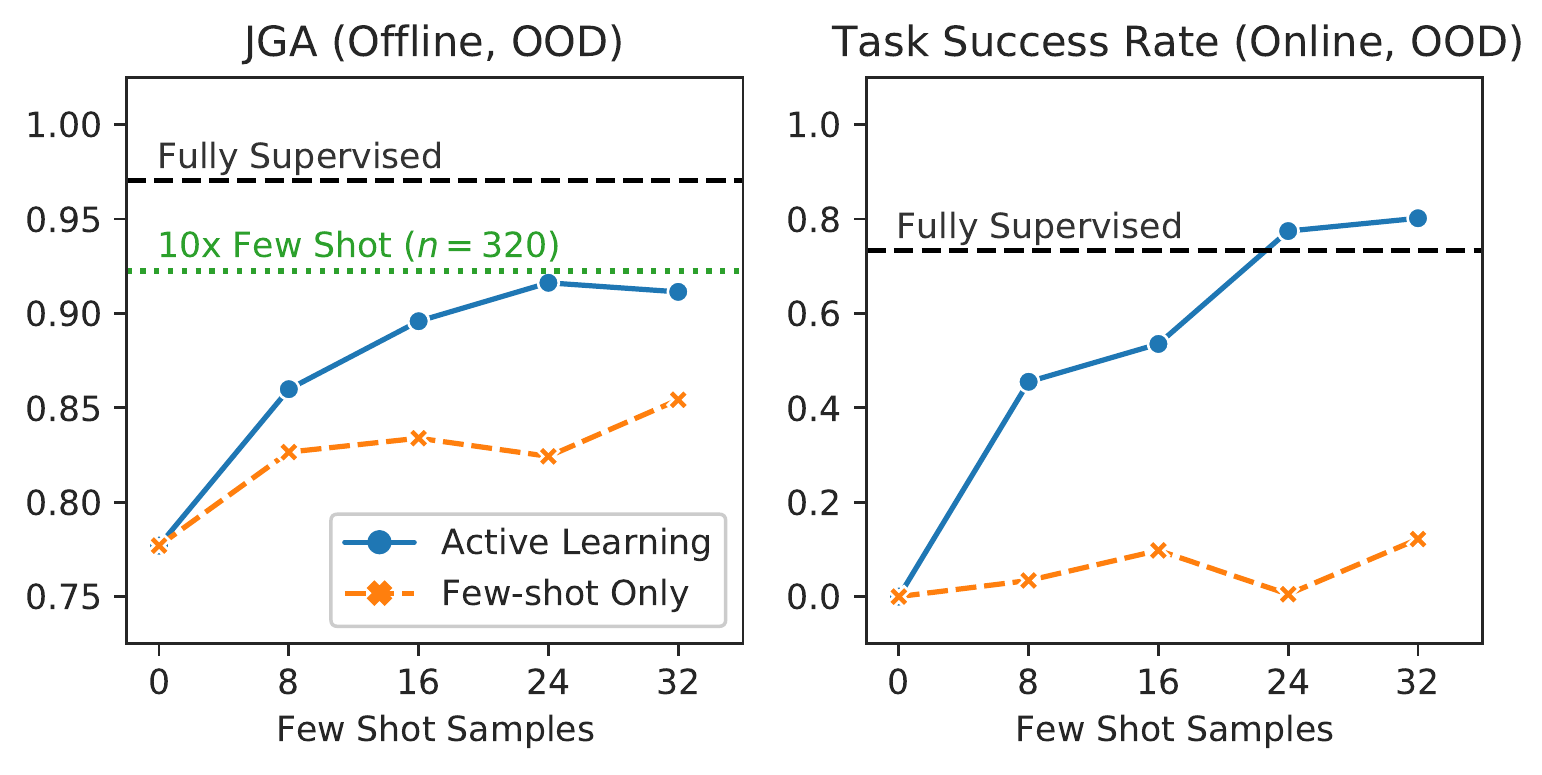}
    \caption{\textbf{Bootstrapped Validation Performance with Active Learning.} Active Learning vastly outperforms a model with an equivalent number of few-shot samples. JGA and TSR match performance of a baseline with $10\times$ more domain-specific samples.}
    \label{fig:activelearning_perf}
\end{figure}

\paragraph{Active Learning}
Results for our Active Learning experiments are shown in Figure~\ref{fig:activelearning_perf}. Contrary to the Zero-shot models, we see that JGA performance consistently improves for 3 iterations, finishing with a total of 13.4 points over the Zero-shot baseline and matching the performance of Few-shot model with 320 dialogues (in green), $10\times$ the number available to the Active Learning model. The Active Learning model also shows a large gain over the sample-equivalent Few-shot model (dashed orange), and that the gain increases with the number of samples. These results demonstrates that our use of Task Success strongly improves our sample-efficiency. TSR performance continues to improve, and eventually exceeds the fully-supervised model that was trained with $72\times$ more data.

Although not shown, we find that Assistant BLEU, as in our Zero-shot experiments similarly does not significantly improve. This indicates that TSR is more strongly correlated with JGA, and its optimization primarily benefits NLU. In-Domain JGA also remains flat, confirming that synthetic data does not lower existing performance. Additional metrics are provided in Appendix~\ref{sec:appextraresults}.

\paragraph{Test Set Results} To ensure that our methodology did not inadvertently overfit via the leaking of goals, we report final Test Set performance for a fully held-out set of Out-of-Domain data; for Simulation-based models, we evaluated models after 4 iterations. Results are shown in Table~\ref{tab:testset}. We find that results are consistent with our earlier analysis. Zero-shot JGA improves 9 points over the baseline, and Active Learning gains 14 points over the baseline. Task Success Rate shows larger improvements, and matches the Fully-supervised baseline. Both models outperform the Few-shot only baseline. Finally, we see that our simulation procedure remains useful even in data-rich environments: adding simulations to a fully supervised model improves JGA by 0.4 absolute points (15\% error reduction).

\begin{table}[t]
    \centering
    \begin{small}
    \begin{tabular}{lrrrr}
    \toprule
    & \multicolumn{2}{c}{{\bf In-Domain}} & \multicolumn{2}{c}{{\bf Out-of-Dom}}\\
    \cmidrule(lr){2-3} \cmidrule{4-5}
    {\bf Model}           &   JGA & TSR &  JGA & TSR\\
    \midrule
    Base model              &  .829 & .394 & .770 & .000\\
    Simulation              &  .838 & .454 & .860 & .779\\
    \mylinesep
    Few-shot only ($n=32$)  &  .835 & .459 & .852 & .140\\
    Active Learning ($n=32$)&  .830 & .362 & .911 & .799\\
    \mylinesep
    Fully Supervised        &  .895 & .551 & .973 & .769\\
    Fully Sup. + Simulation &  .895 & .555 & .977 & .847\\
    \bottomrule
    \end{tabular}
    \end{small}
    \caption{\textbf{Test set results.} Final performance on the
    held out Test-set for both In-Domain and Out-of-Domain.}
    \label{tab:testset}
\end{table}

\paragraph{Human Evaluation}

We perform a final human evaluation using each the models from our experimental conditions. We repeat the ACUTE-Evals described in Section~\ref{sec:synthdataresults}, using synthetic dialogues from each condition. Results are shown in Figure~\ref{fig:acute_metrics_final}.

We find that all human judgements are roughly consistent with our expectations from offline evaluation, with clear wins for Active Learning over the Baseline and Few-shot models. In all cases, the Human gold data significantly outperforms all of our models, indicating further avenues for improvement. Nonetheless, the win rate of Humans decreases in our models compared to the Baseline and Few-shot models.

In analysis of annotator preferences, we find that annotator selection is generally well-correlated with TSR: Annotators preferred unsuccessful conversations over successful conversations in only about 10\% of pairings. Simplicity and clarity were oftentimes given as rationale for preference in these pairings: annotators generally preferred Assistants that had clear communication. Many of the models learned to ask confirmation questions; this was generally liked by the Annotators as long as there was not too much back and forth between the User and Assistant models in doing so. Some Annotators even preferred generated conversations with confirmations over the gold, human conversations. While some Annotators preferred ``friendlier'' or ``more conversational'' Assistants, this was not a consistent preference; some Annotators found similar conversations to be ``weirdly informal''
or to ``take too long to get to the point.''

\begin{figure}[t]
\centering
\begin{small}
\begin{tabular}{lr|rrrrr}
\multicolumn{2}{c}{} & \multicolumn{5}{c}{Win \%}\\
\multicolumn{2}{c}{} & B & F & Z & A & H\\
\cmidrule(lr){3-7}
\parbox[t]{2mm}{\multirow{4}{*}{\rotatebox[origin=c]{90}{Lose \%}}}
 & \underline{B}ase Model           &              & \POS{\bf.75} & \POS{\bf.74} & \POS{\bf.74} & \POS{\bf.91} \\
 & \underline{F}ew-shot only        &  \POS{\bf.25} &              & \POS{\bf.75} & \POS{\bf.67} & \POS{\bf.71}\\
 & \underline{Z}ero-Shot Sim.       & \POS{\bf.26} & \POS{\bf.25} &              & \NEG{.46}    & \POS{.58}   \\
 & \underline{A}ctive Learning      & \POS{\bf.26} & \POS{\bf.33} & \NEG{.54}    &              & \POS{\bf.69}\\
 & \underline{H}uman                & \POS{\bf.09} & \POS{\bf.29} & \POS{.42}    & \POS{\bf.31} &             \\
\end{tabular}
\end{small}
\caption{\textbf{Pairwise Human Evaluations of bootstrapped models}.
Entries marked in blue agree with automatic metrics, while those in red italics disagree with
the automatic metrics. Bold numbers indicate statistical significance ($p < .05$, binomial test).
}
\label{fig:acute_metrics_final}
\end{figure}

\section{Limitations}

\begin{figure}[t]
    \centering
    \includegraphics[width=0.99\linewidth]{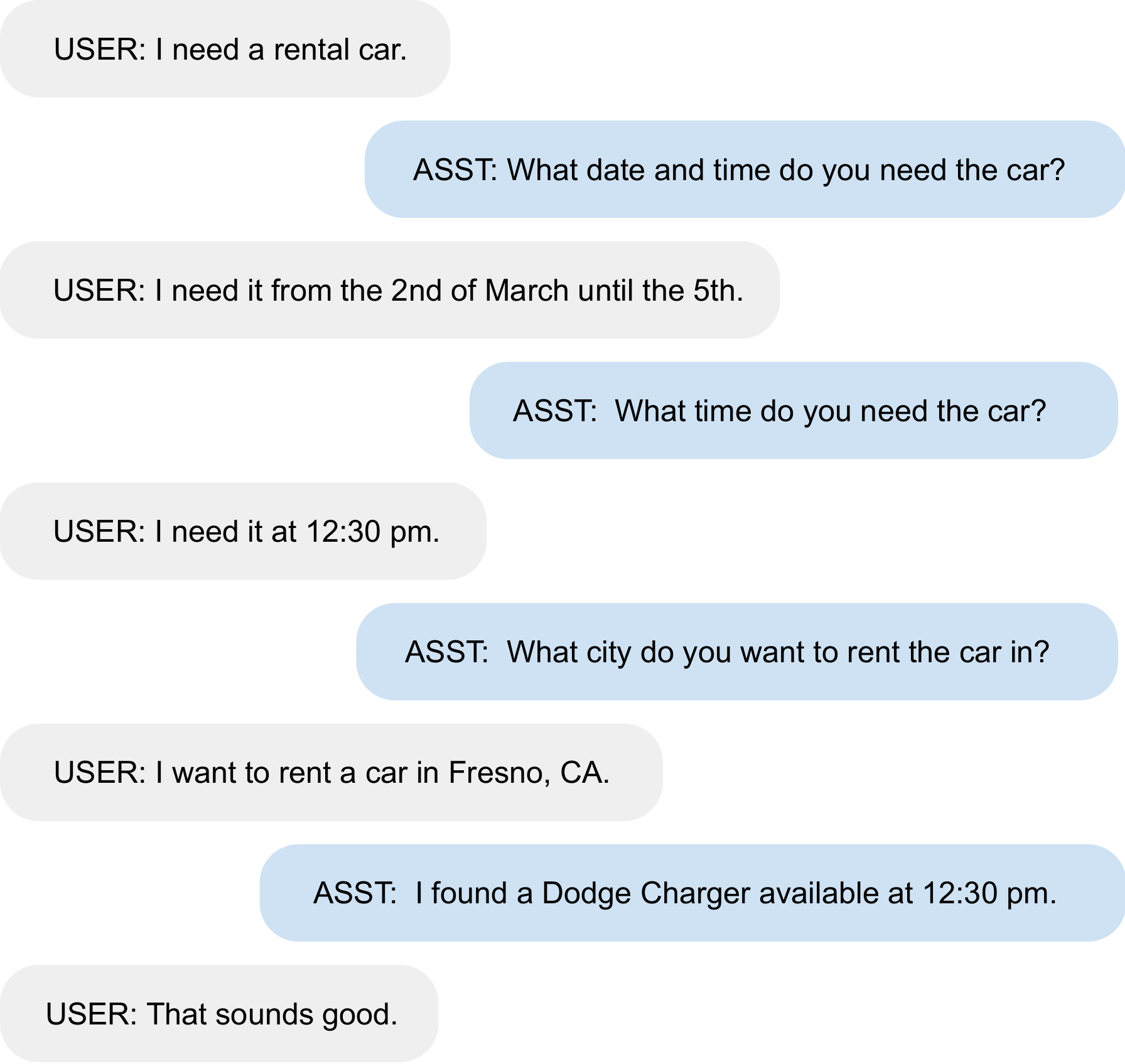}
    \caption{\textbf{Example Synthetic Conversation} generated by our final Active Learning model, using \textit{Greedy} generation. Although the conversation is successful, the linguistic variation is low.}
    \label{fig:greedy_gen}
\end{figure}

While our bootstrapping and Active Learning procedures do significantly improve the robustness of models, they do surprisingly little to affect linguistic diversity, especially when using greedy generation. As a representative example, see Figure~\ref{fig:greedy_gen}. In it, we observe that the conversation devolves to a simple slot-filling questionnaire, with the User beginning many utterances with ``I need\ldots'' Manually reviewing simulated conversations found that while hallucination is very low in greedy-generated dialogues, most dialogues form roughly the slot-filling questionnaire pattern. While the synthetic conversations are plausible, their linguistic diversity is extremely low, explaining why our Task Success Rates reach near perfect levels: the User learns to specify things as simply as possible. Furthermore, we find one of our domains (\textit{Make payments}) has multiple instances of infinite-loops being generated, a common issue known in Neural Language Models \cite{holtzman2020nucleus,welleck2019neural}. 

\begin{figure}[t]
    \centering
    \includegraphics[width=0.99\linewidth]{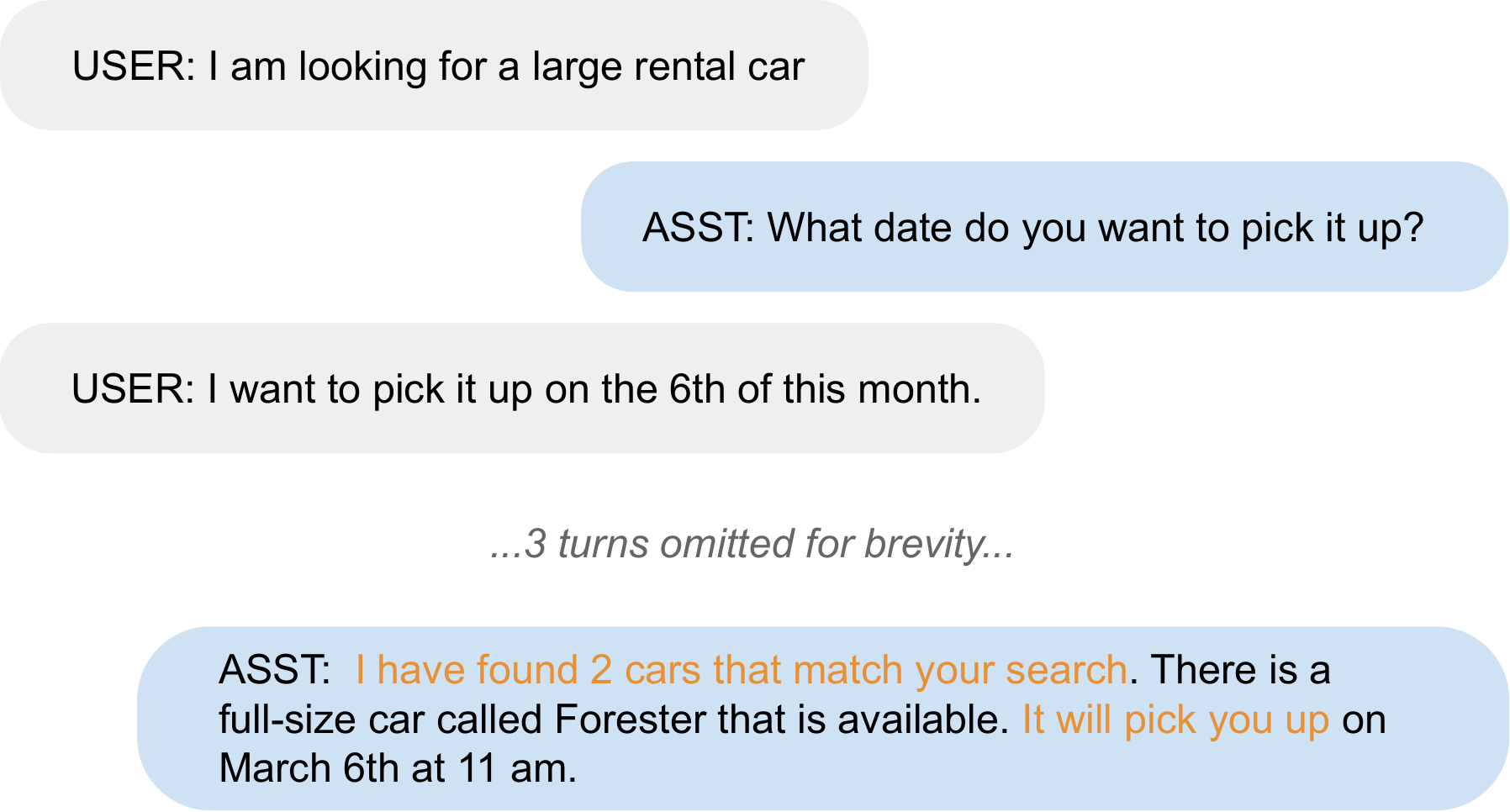}
    \caption{\textbf{Example Synthetic Conversation} using \textit{Nucleus} generation. Linguistic diversity is improved, but the Assistant hallucinates information (in orange).}
    \label{fig:nucleus_gen}
\end{figure}

We also show an excerpt from a similar dialogue except with Nucleus sampling in Figure~\ref{fig:nucleus_gen}. This excerpt from the \textit{Rental Cars} domain is mostly representative of the synthetic data generated for our training runs. We see increased linguistic variation, but we also observe that the Assistant model hallucinates: the API response does not say how many cars are available (it only provides one), and rental cars are unlikely to pick up their clients (likely learned from the \textit{Ride Share} domain). Upon further inspection, we discovered this problem was ubiquitous for the Base model, but only present in limited quantities on bootstrapped models -- suggesting retraining actually \textit{improves} this behavior.

These examples, along with the lack of improvements in NLG metrics (BLEU scores), show that Task Success is likely to reduce the linguistic diversity of Assistants or Users, unless generation methods prone to hallucination are used. In the future, identifying automatic-filtering techniques for NLG utterances, similar to Task Success and slot filling, could help with this problem. Other generation methods, like Diverse Beam Search \cite{vijayakumar2016diverse}, may also be able to perform a better hallucination-diversity trade-off. Alternatively, the model could only be trained on the API Call turns of synthetic data. Nonetheless, the offline metrics on the Test set demonstrate that our methodology does improves robustness of the Natural Language Understanding components of our model, as indicated by the Out-of-Domain JGA metrics.

\section{Conclusion}

We explored the use of pretrained Language Models as User Simulators in order to generate synthetic dialogues, and filter these models for quality using Task Success. We demonstrated our methodology can be used to improve models in zero-shot, few-shot, and full-shot manners. By incorporating Active Learning, we additionally show that our models are able to bootstrap NLU performance to that of a model with $10\times$ more training data. We encourage future work to look for improved generation methods which improve diversity without hallucination, and to find methods for automatically grading the quality of generations. Other improvements, such as the use of Schema Descriptions \cite{rastogi2019towards,lin2021zero}, may provide further generalization on unseen domains.

\section*{Acknowledgements}
Thank you to members of the Facebook dialogue teams for their helpful feedback and suggestions on this project. We are particularly grateful to Justin Cho, Jianguo Zhang, Shahin Shayandeh, Ahmad Beirami, and Arthur Szlam for their particularly detailed discussions. We also thank Weiyan Shi and Jesse Thomason for their feedback on early drafts of this paper.

\bibliography{anthology,custom}
\bibliographystyle{acl_natbib}

\clearpage

\onecolumn
\appendix

\section{Appendix}
\label{sec:appendix}

\subsection{Screenshot of Annotator UI}

\label{sec:screenshot}
\begin{figure}[h]
    \centering
    \includegraphics[width=\linewidth]{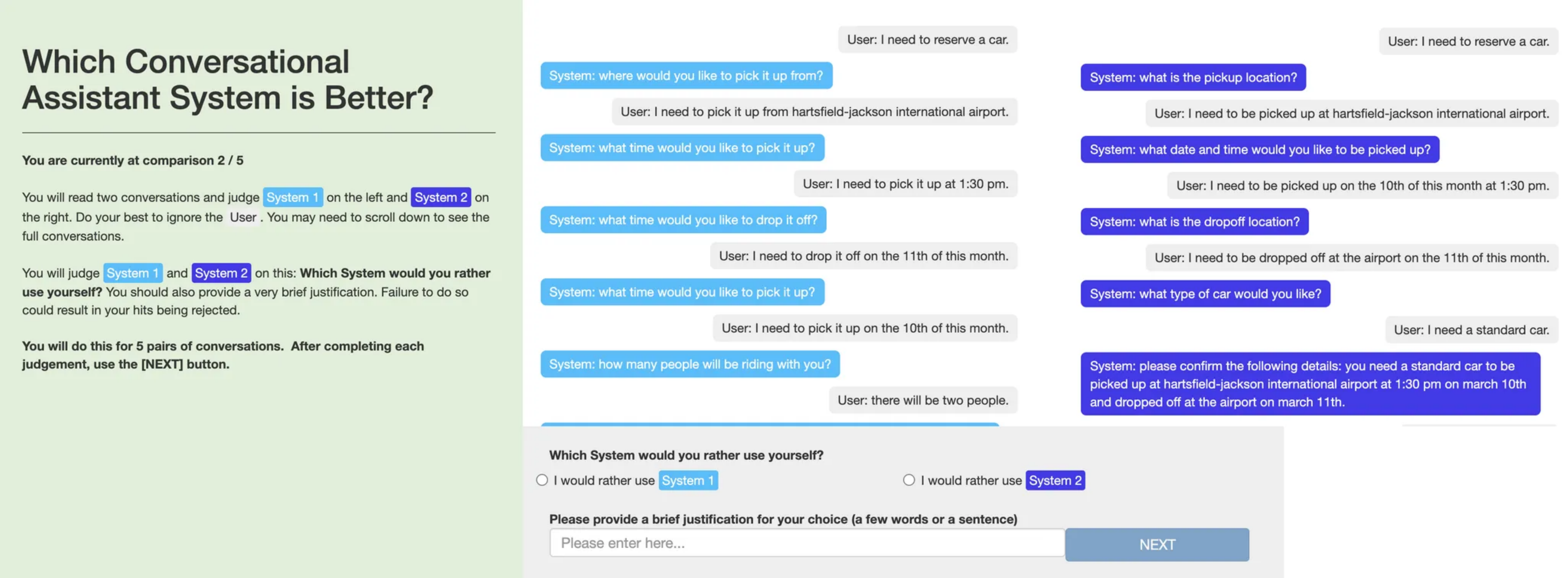}
    \caption{\textbf{Screenshot of Annotator UI}. Annotators are asked to evaluate "Which Conversational Assistant System is better" by pressing radio buttons corresponding to two presented conversations. }
    \label{fig:annotatorUi}
\end{figure}

\subsection{Pretraining Datasets}
\label{sec:appdatasets}

\begin{table}[h!]
\centering
    \begin{tabular}{m{3cm}rrm{9cm}}
    \toprule
    {\bf Dataset}     & {\bf Dial.} & {\bf Dom.} & {\bf Overlap with Google SGD} \\
    \midrule
    GoogleSGD In-Domain     & 18,986 & 16 & Alarm, Banks, Buses, Calendar, Events, Flights, Hotels, Media, Movies, Music, Restaurants, Rideshare, Services, Travel, Trains, Weather\\
    GoogleSGD Out-of-Domain &  3,839 & 4 & Home Search, Messaging, Payment, Rental Cars\\
    \midrule
    MetaLWoZ                & 37,884 & 47 & Banks, Buses, Events, Movies, Music, Restaurants\\
    MSR-E2E                 & 10,087 &  3 & Movies, Restaurants, Taxis\\
    MultiDoGo               & 19,522 &  6 & Calendar, Flights, Media, Weather\\
    MultiWoz                & 10,438 &  7 & Attractions, Hospitals, Hotels, Restaurants, Taxis, Train\\
    Taskmaster-1            & 13,215 &  6 & Restaurants, Rideshare\\ 
    Taskmaster-2            & 17,289 &  7 & Flights, Hotels, Movies, Music, Restaurants\\
    TicketTalk              & 23,789 &  1 & Movies\\
    \bottomrule
    \end{tabular}
    \caption{\textbf{Detailed statistics of datasets used in our work}. All datasets except for Google SGD Out-of-Domain used in pretraining.}
    \label{tab:detailedDatasets}
\end{table}

We describe the different datasets that we use for pretraining. In general, we attempt to make these datasets be structured as similarly as possible to the conversations format as described in Sec \ref{sec:neuralusersim} and generate data for separate User and Assistant models once formatted. For datasets with no API Call or Response labels, we imitate these values by accumulating dialogue state across user and assistant responses, respectively, and presenting these on appropriate turns. Other exceptions are described inline. 

See Table \ref{tab:detailedDatasets} for dataset statistics. 

\paragraph{Google SGD} We describe Google SGD in \ref{googleSgdDescripition}. We describe our method for splitting Google SGD into In-Domain and Out-of-Domain splits in \ref{sec:bootstrapexp}. 

For our Out-of-Domain split, we used Home Search, Messaging, Payment, Rental Cars as 4 holdout domains that did not have analogues in any of the other datasets. Though the "Services" and "Travel" domains do not occur explicitly in the other datasets, we do not include them in our holdout since they include semantically similar information to the "Hospital" and "Attractions" domains of MultiWoz, respectively. For our In-Domain split, we include solely the 16 other domains of the dataset. 

As also mentioned in \ref{sec:bootstrapexp}, since Google SGD is a dataset that contains both single-goal and mulit-goal conversations, some domains of the In-Domain split are present as goals in multi-goal conversations of the Out-of-Domain split.

\paragraph{MetalWoz}

MetalWoz is a dataset constructed in a Wizard of Oz fashion across 227 tasks and 47 domains. Given a domain and a task, conversing pairs were asked to chat for 10 turns to satisfy the user’s queries.

As this dataset does not include any annotations about API Calls, API Responses, or belief state, we pretrain on this dataset as-is and do not attempt to transform it into the format described in Sec \ref{sec:neuralusersim}. We do however split this dataset into separate User and Assistant versions.

\paragraph{MultiDoGo}
MultiDoGo is a large task-oriented dataset collected in a Wizard of Oz fashion, using both crowd and expert annotators with annotations at varying levels of granularity. We use only the data available publicly on this dataset's open-source repository (about 20k dialogues total.)

\paragraph{MultiWoz} 
MultiWoz is a dataset of single and multi-goal human-human conversations collected in a Wizard of Oz fashion. Validation and test sets contain only successful conversations while the train set include some that are incomplete. Data of the original dataset is labelled with belief states. 

\paragraph{MSR-E2E} 
MSR-E2E is a dataset of human-human conversations in which one human plays the role of an Agent and the other one plays the role of a User. Data is collected from Amazon Mechanical Turk. 

\paragraph{Taskmaster 1}
Conversations in Taskmaster 1 were collected in one of two ways: spoken Wizard of Oz conversations between humans (transcribed to text) as well as written conversations from a single human in a self-dialog method. Similar to our conversations format, rather than being labelled with intents and dialog acts, conversations of this dataset are labelled with simple API arguments.

\paragraph{Taskmaster 2}
Taskmaster2 is a dataset of entirely spoken two person dialogues collected in a Wizard of Oz manner where Assistant utterances were typed by a human and then "spoken" using a text-to-speech service. Dialogues in this dataset includes those that are search and recommendations oriented, rather than purely task execution.

\paragraph{TicketTalk}
TicketTalk (or Taskmaster 3) is a dataset of movie ticket dialogues collected in a self-chat manner. To induce conversational variety, crowd workers were asked to generate conversations given dozens of different instructions of different level of specificity, some purposefully including conversational errors.

\clearpage
\subsection{Hyperparameter Tables}

We include hyperparameter tables for each of our models used in this paper. All models were trained using the ADAM optimizer. All models were optimized using Token Exact Match (examples with perfect greedy decoding) as an early stopping criteria. 

\paragraph{Evaluating Synthetic Dialog Generation and its Metrics}
Note that as described in Sec \ref{sec:synthetic}, we aim to look for reasonable correlations between metrics and model architectures rather than absolute performance.

\paragraph{LSTM \& LSTM with Attention:} (1 GPU per run)

\begin{center}
\begin{small}
\begin{tabular}{lr}
    \toprule
    {\bf Hyperparameter} & {\bf Swept Values} \\
    \midrule
    Learning Rate & 1e\{-3, -4\}\\
    Number of Layers & \{1, 2, 4\}\\
    Embedding size & \{256, 384\}\\
    Hidden size & \{1024, 2048\}\\
    Batch size & 64\\
    Embedding Init & FastText\\
    \bottomrule
\end{tabular}
\end{small}
\end{center}

\paragraph{GPT2:} (8 GPUs per run)
\begin{center}
\begin{small}
\begin{tabular}{lr}
    \toprule
    {\bf Hyperparameter} & {\bf Swept Values} \\
    \midrule     
    Learning Rate & 1e{\{-5, -6\}}\\
    LR Scheduler & Reduce on Plateau, Invsqrt \\
    Model Size & 124M \\
    Text Truncate & 512 \\
    Warm-up Updates & 100 \\ 
    Batch size & 4 \\
    Update Frequency & 2 \\
    Gradient Clip & 1 \\
    \bottomrule
\end{tabular}
\end{small}
\end{center}

\paragraph{BART:} (8 GPUs per run)
\begin{center}
\begin{small}
\begin{tabular}{lr}
    \toprule
    {\bf Hyperparameter} & {\bf Swept Values} \\
    \midrule     
    Learning Rate & 1e{\{-5, -6\}}\\
    LR Scheduler &  \{Reduce on Plateau, Invsqrt\} \\
    Model Size & 400M \\
    Text Truncate & 512 \\
    Warm-up Updates & 100 \\ 
    Batch size & 4 \\
    Update Frequency & 2 \\
    Gradient Clip & 1 \\
    \bottomrule
\end{tabular}
\end{small}
\end{center}

\paragraph{Bootstrapping Novel Domains} 

Once we early stopped models on Token Exact Match, we used TSR on validation goals of Google SGD to select the best model out of a given hyperparameter sweep. However, a post hoc analysis suggests that Token Exact Match would have worked approximately as well for the goal of improving JGA. 

\paragraph{BART:} (8 GPUs per run)
\begin{center}
\begin{tabular}{lr}
    \toprule
    {\bf Hyperparameter} & {\bf Swept Values} \\
    \midrule     
    Learning Rate & \{1e-4, 5e-5, 1e-5, 5e-6\}\\
    Model Size & 400M\\
    Batch size & 4 \\
    Update frequency & 8 \\
    LR Scheduler & Invsqrt \\
    Warm-up updates & 1000 \\
    Text truncate & 512 \\
    Label truncate & 512 \\
    Gradient Clip & 0.1 \\
    Multitask Weights & 1 \\
    Validation Steps & 100 \\
    \bottomrule
\end{tabular}
\end{center}

\clearpage

\subsection{Additional Results}
\label{sec:appextraresults}

We report additional metrics on each of our models, for both offline (static, held-out data) and online (during simulation) settings.

\begin{figure}[h]
    \centering
    \includegraphics[width=\linewidth]{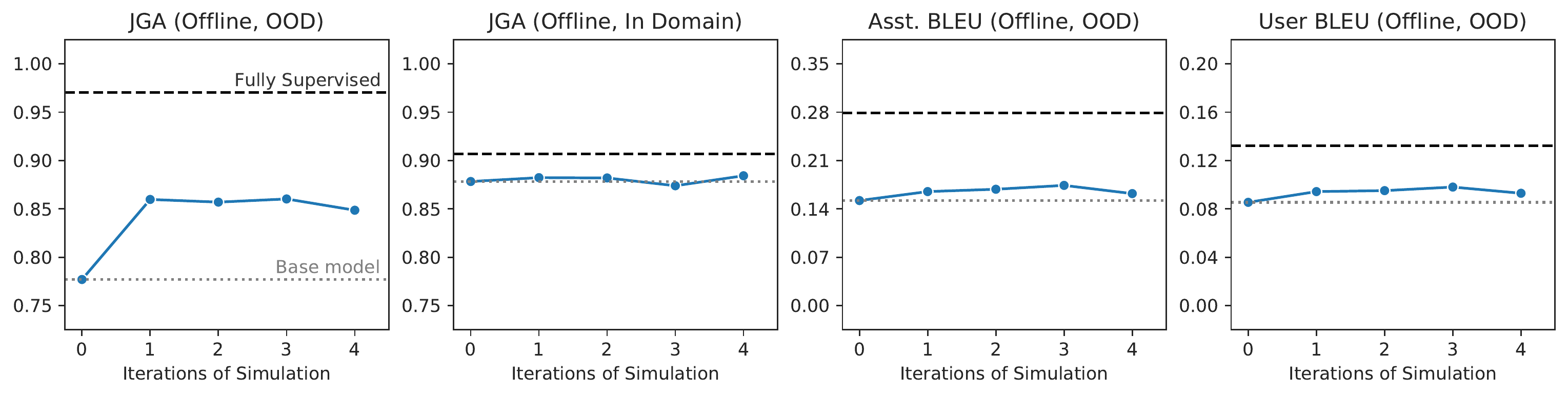}
    \caption{{\bf Offline Bootstrapping Results.} Results of Bootstrapping on a static (held-out) offline dataset.}
    \label{fig:inrl_offline_results}
\end{figure}

\begin{figure}[h!]
    \centering
    \includegraphics[width=\linewidth]{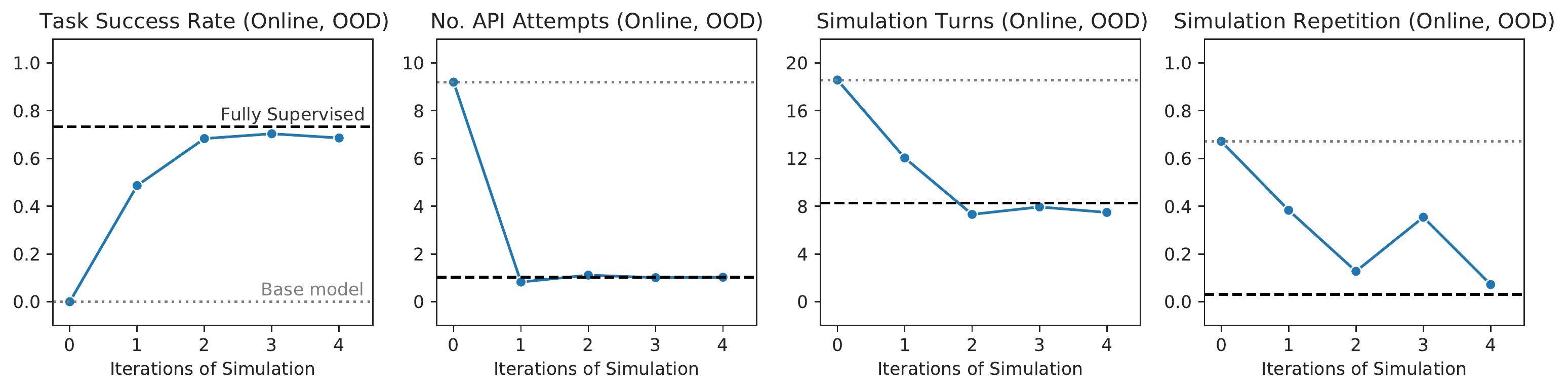}
    \caption{{\bf Online Bootstrapping Results.} Results of Bootstrapping models on during simulations.}
    \label{fig:inrl_online_results}
\end{figure}

\begin{figure}[h!]
    \centering
    \includegraphics[width=\linewidth]{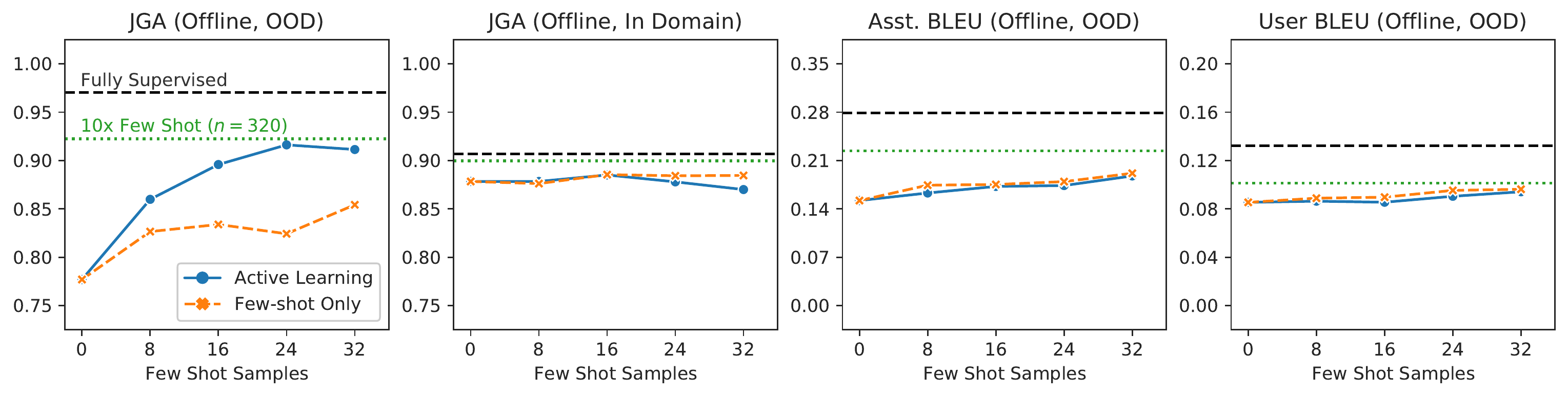}
    \caption{{\bf Offline Active Learning Results.} Results of Active Learning on a static (held-out) offline dataset.}
    \label{fig:al_offline_results}
\end{figure}

\begin{figure}[h!]
    \centering
    \includegraphics[width=\linewidth]{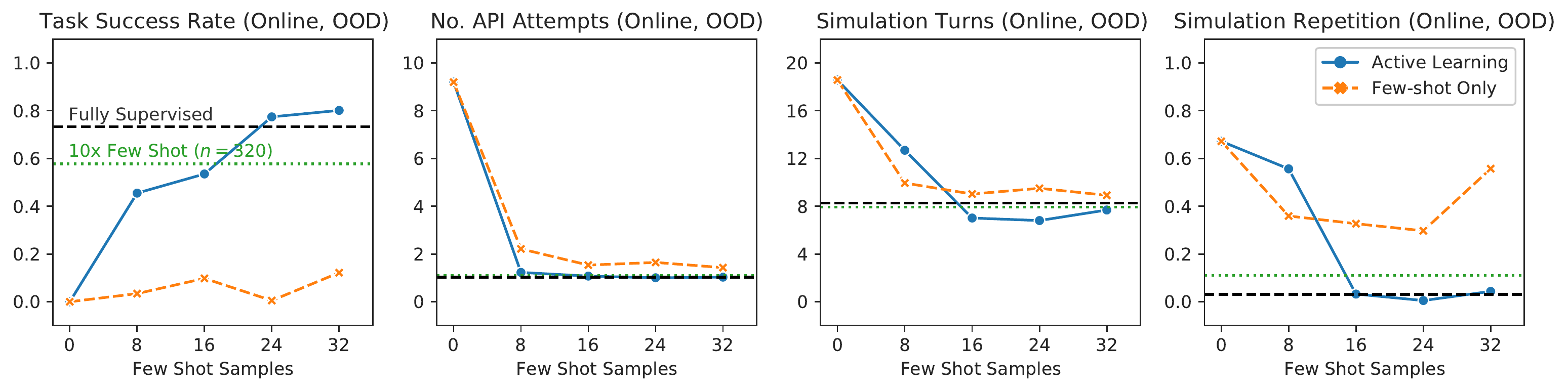}
    \caption{{\bf Online Active Learning Results.} Results of Active Learning model during simulations.}
    \label{fig:al_online_results}
\end{figure}

\clearpage

\subsection{Holdout API Analysis of Bootstrap Procedure}
\label{sec:deeperAnalysis}

We take the models generated from Sec \ref{sec:bootstrapexp} and evaluate the models for JGA, limiting only turns to which include API calls, over each of the holdout domains. Results of this are shown in Figure~\ref{fig:deeperAnalysisJGA}. We observe that all holdout domains have a JGA value of zero for the Base model; this validates our selection of holdout domains. We also observe that compared to the other models, Active Learning performs much better across all domains.

\begin{figure}[h!]
\centering
\begin{small}
\begin{tabular}{l cccc}
\toprule
\multicolumn{1}{c}{} & \multicolumn{4}{c}{\bf Domain}\\
\cmidrule(lr){2-5}

\bf{Model} & Find Homes & Payment & Rental Cars & Messaging \\
\toprule
  Base Model           &  .000            & .000  & .000  & .000   \\
  Few-shot only        &  .603 &      .022        & .411 & .768 \\
  Zero-Shot Sim.       & .876 & .022 &        .266      & .929   \\
  Active Learning      & \bf{.882} & \bf{.870} & \bf{.623}   &      \bf{.946}       \\
\bottomrule
\end{tabular}
\end{small}
\caption{\textbf{JGA of individual Holdout Domains}  (limited to API Call Turns only).}
\label{fig:deeperAnalysisJGA}
\end{figure}

\end{document}